\documentclass[sigconf, screen, dvipsnames, authorversion]{acmart}

\usepackage{hyperref}
\usepackage[noend]{algorithmic}
\usepackage{algorithm}
\usepackage{multirow}
\usepackage{amsmath}
\usepackage{bbm}
\usepackage{booktabs}
\usepackage[inline]{enumitem} 
\usepackage{subcaption}
\usepackage{bm} 
\usepackage{xcolor}

\usepackage{lipsum}

\usepackage{wrapfig}
\usepackage[hyphenbreaks]{breakurl}

\usepackage{arydshln}
\makeatletter
\def\adl@drawiv#1#2#3{
        \hskip.5\tabcolsep
        \xleaders#3{#2.5\@tempdimb #1{1}#2.5\@tempdimb}%
                #2\z@ plus1fil minus1fil\relax
        \hskip.5\tabcolsep}
\newcommand{\cdashlinelr}[1]{%
  \noalign{\vskip\aboverulesep
          \global\let\@dashdrawstore\adl@draw
          \global\let\ adl@draw\adl@drawiv}
  \cdashline{#1}
  \noalign{\global\let\adl@draw\@dashdrawstore
          \vskip\belowrulesep}}
\makeatother

\usepackage{tikz}
\usetikzlibrary{automata, arrows, bayesnet, bending}

\usepackage{tikzscale}

\copyrightyear{2025}
\acmYear{2025}
\setcopyright{acmlicensed}
\acmConference[UMAP Adjunct '25]{Adjunct Proceedings of the 33rd ACM Conference on User Modeling, Adaptation and Personalization}{June 16--19, 2025}{New York City, NY, USA}
\acmBooktitle{Adjunct Proceedings of the 33rd ACM Conference on User Modeling, Adaptation and Personalization (UMAP Adjunct '25), June 16--19, 2025, New York City, NY, USA}
\acmDOI{10.1145/3708319.3734172}
\acmISBN{979-8-4007-1399-6/2025/06}

\begin{document}
\title{Procedural Memory Is Not All You Need}
\subtitle{Bridging Cognitive Gaps in LLM-Based Agents}
\author{Schaun Wheeler}
\affiliation{
  \institution{aampe}
  \state{North Carolina}
  \country{USA}
}
\email{schaun@aampe.com}
\author{Olivier Jeunen}
\affiliation{
  \institution{aampe}
  \city{Antwerp}
  \country{Belgium}
}
\email{olivier@aampe.com}

\begin{abstract}
Large Language Models (LLMs) represent a landmark achievement in Artificial Intelligence (AI), demonstrating unprecedented proficiency in procedural tasks such as text generation, code completion, and conversational coherence. These capabilities stem from their architecture, which mirrors human \textit{procedural memory}---the brain’s ability to automate repetitive, pattern-driven tasks through practice.
However, as LLMs are increasingly deployed in real-world applications, it becomes impossible to ignore their limitations operating in complex, unpredictable environments.
This paper argues that LLMs, while transformative, are fundamentally constrained by their reliance on procedural memory.
To create agents capable of navigating ``wicked'' learning environments---where rules shift, feedback is ambiguous, and novelty is the norm---we must augment LLMs with semantic memory and associative learning systems.
By adopting a modular architecture that decouples these cognitive functions, we can bridge the gap between narrow procedural expertise and the adaptive intelligence required for real-world problem-solving.  
\end{abstract}

\maketitle

\section{Cognitive Limitations of LLMs in Autonomous Decision-Making}
Large language models excel at generating fluent procedural outputs but often falter when faced with dynamic, ``wicked'' environments that require flexible reasoning and memory recall. In this paper, we first provide a cognitive‐science‐informed analysis of why procedural‐memory‐centric LLM architectures fail in complex tasks, and propose a modular system that augments LLMs with dedicated semantic and associative memory components to support causal decision‐making. We then review the limitations of procedural memory in LLMs, detail our three‐part modular architecture, and compare our approach to related work and illustrate its benefits.
\subsection{LLM Architecture}
LLMs operate as agentic actors through architectures rooted in transformer-based sequence modeling. Their core mechanism---pro\-cedural memory---is implemented via self-attention layers that statistically model token co-occurrence patterns across massive text corpora~\cite{Vaswani2017}.

LLMs generate outputs by computing attention weights across input tokens. This mechanism captures \textit{local} and \textit{global} dependencies, enabling probabilistic predictions of the next tokens. Similarly to procedural memory in the basal ganglia, which strengthens synaptic connections through repeated task execution (e.g., piano practice)~\cite{Foerde2011}, LLMs refine attention weights during training to automate pattern completion.  

LLMs often return factually incorrect information because their procedural memory---trained to predict sequences by weighting token co-occurrences---generates outputs based on statistical likelihoods in their training data, not grounded truth~\cite{Ruis2025}.
The noise between linguistic patterns and real-world facts creates a gap between the model's plausible token continuations and verified knowledge.
``Hallucinations'' or ``confabulations'' (fluent but ungrounded token continuations~\cite{Smith2023}), thus emerge when statistical priors override factual precision. 

In short, LLMs lack episodic memory. 
In an attempt to combat this, Retrieval-Augmented Generation (RAG) systems combine a dense retriever with an LLM~\cite{Lewis2020}, conditioning the LLM's output on both a query \emph{and} and a document.
Whilst a step in the right direction, this architecture remains fundamentally limited.

\subsection{Limitations of Procedural Memory}
LLM architecture, even when augmented by RAG systems, exhibits several critical limitations:
\begin{itemize}
    \item LLMs operate without persistent state, meaning they cannot retain information from previous interactions unless that information is explicitly included in the input prompt. Each inference is an independent forward pass, and prior interactions are not retained unless explicitly re-injected via prompts. 
    \item LLMs are constrained by fixed-length context windows (typically a few thousand to a few hundred-thousand tokens), which limit their ability to process or recall long sequences of information across time. Even models with large windows truncate or discard prior context~\cite{Hosseini2025}, violating the continuity required for episodic reasoning (the ability to recall and integrate specific past experiences, including their temporal and contextual details, to inform present decisions~\cite{Maurer_Nadel_2021}).
    \item LLMs do not have a memory-consolidation mechanism to integrate and retain learned experiences over time. Humans consolidate episodic memories into long-term storage via hippocampal replay~\cite{Olafsdottir2018}. More recent developments have introduced mechanisms for storing summaries of past interactions to simulate memory consolidation in LLMs~\cite{zhong2023}, but these approaches capture only surface-level recaps rather than integrating experiences into structured knowledge or enabling flexible recall based on relevance or context.
    \item Even with RAG, retrieved documents are injected as static context tokens~\cite{Gao2024}. Unlike biological episodic memory, which dynamically updates and re-weights past experiences, RAG cannot revise retrieved knowledge mid-interaction (e.g., reconciling conflicting facts in documents).  
    \item LLMs have no mechanism for meta-learning. During inference, an LLM relies on fixed parameters, processing retrieved data as context without modifying its attention mechanisms~\cite{Liu2025}. Unlike humans, it cannot reshape attention based on new insights, which limits real-time adaptability.
\end{itemize}
A customer service LLM exemplifies these limitations. Since it is stateless, it forgets past interactions unless the user explicitly repeats details or the system re-injects them into the prompt. Its fixed context window means that prior messages may be truncated, preventing seamless episodic reasoning. Unlike human memory, it cannot consolidate key details---such as a user's repeated complaints about delayed shipping---into long-term storage. Even with RAG, retrieved records are static and cannot be updated mid-session, meaning that conflicting information (e.g., an initial delay estimate vs. a later update) is not reconciled dynamically. Finally, because it lacks meta-learning, the LLM does not refine how it processes such interactions over time, forcing users to repeatedly provide the same context rather than benefiting from accumulated experience.

It could be argued that fine-tuning based on user histories could simulate episodic memory. However, fine-tuning introduces catastrophic forgetting: updating weights for new data degrades performance on prior tasks~\cite{Goodfellow2015}.
Biological episodic memory avoids this via neurogenesis and sparse coding~\cite{Wixted2014}, which transformers do not replicate.

All of these challenges are not specific to LLMs. They are specific to procedural memory. The flaw is not in the implementation of LLMs, but in their architectural underpinnings.
\subsection{The Need for Non-Procedural Memory}
In addition to the inherent limitations of an architecture that is limited to procedural memory (plus a problematic implementation of episodic memory), LLMs also exhibit semantic memory gaps. LLMs encode knowledge as dense, overlapping vector representations in high-dimensional latent spaces. Although these embeddings capture statistical relationships (e.g., ``Paris is to France as Tokyo is to Japan''), they lack explicit hierarchies (no discrete nodes for facts---e.g., ``User $X$ prefers eco-friendly products''), and symbolic grounding (vectors conflate syntactic, semantic, and pragmatic features). 

While embeddings do form clusters (e.g., animals vs. vehicles), they cannot perform compositional reasoning (e.g., inferring ``If all mammals breathe air, and whales are mammals, then whales breathe air'') probabilistically. LLMs generalize logical structures by recognizing patterns from training data. This falls short of true semantic reasoning, and therefore LLMs often fail in situations that require strict logical consistency.
LLMs also exhibit systematic associative blind spots. Transformers model pairwise token interactions but struggle with higher-order associations that require chained reasoning or cross-context links~\cite{Peng2024}.

For example, associating a user's preference for ``transparency'' (from a shipping complaint) with ``detailed ingredient list'' (in a food app) requires inferring abstract principles, not token co-occurrence. This is an architectural constraint, not a limitation in training or data.
Attention heads focus on local context, limiting cross-session reasoning. Whereas humans link concepts via bidirectional hippo\-campal-cortical pathways, transformers process information uni-directionally (input leads to output, not the other way around).  

It could be argued that chain-of-thought prompting enables multi-step reasoning~\cite{Wei2022}, but chain-of-thought relies on \textit{procedural} pattern extension, not associative binding. It cannot dynamically link concepts learned in separate sessions (e.g., connecting a user's travel preferences to their shopping habits). While advancements like elastic weight consolidation (which preserves important parameters across tasks to reduce forgetting) and dynamic context window management (e.g., recurrent memory transformers that extend context beyond fixed token limits) have made strides in addressing some limitations of LLMs, they remain constrained by fundamental architectural mismatches. Elastic weight consolidation has been shown to mitigate catastrophic forgetting in ``kind'' learning environments (Atari games)~\cite{Kirkpatrick2017}, but it has yet to be shown how well this method performs in continuous, non-stationary learning, where feedback loops are sparse and ambiguous.

Similarly, while dynamic context windows improve memory retention~\cite{Wang2024}, they remain bound to transformers' procedural framework, and are therefore limited by that framework's general architectural constraints---processing tokens rather than discrete concepts, and lacking interfaces to ground semantics or associate rewards with specific knowledge structures.

The question is not merely whether LLMs can be extended to overcome the absence of dynamic memory systems, but whether it is more realistic to adapt an architecture to do something it was not designed to do---forcing transformers to mimic episodic and semantic memory---or to append an architecture explicitly designed for these functions. Appending new components can create brittle reliance on prompt engineering or retrieval heuristics. Re-thinking the architecture allows for more principled interaction patterns and targeted learning.

\subsection{The Need for Modular AI}
AI agents do not need to ``think'' exactly the way humans do in order to be useful. In fact, biological systems evolved through constraints (e.g., energy efficiency)~\cite{Li2020}, not optimal design, so trying to replicate human cognitive systems exactly might be a suboptimal approach.
At any rate, modular AI architectures are pragmatic engineering choices to compensate for transformer limitations, not attempts to replicate biology.
LLMs' procedural prowess is undeniable, but their architectural rigidity---static parameters, dense embeddings, and unidirectional processing---renders them inadequate for decision-making in \emph{unkind} or ``\emph{wicked}'' environments.

While techniques like fine-tuning, RAG, and chain-of-thought mitigate specific issues, they fail to address the core absence of dynamic memory systems. The path forward lies in hybrid architectures that pair LLMs with associative and semantic modules, explicitly designed for incremental learning and cross-context reasoning.  

\section{Learning Environments}
The degree to which an agent is capable of successfully acting with autonomy is inextricably tied to the structure of the learning environment in which it operates. Learning environments are a concept rooted in cognitive science and decision-making research. Psychologist Robin Hogarth's framework of ``kind'' and ``wicked'' learning environments has important implications for agent design~\cite{Hogarth2015}. In Hogarth's original formulation of learning environments, \textit{kind} environments are defined by stable rules, repetitive patterns, and unambiguous feedback. Examples include chess, standardized testing, and cooking from a recipe. In such environments, outcomes are immediate and directly attributable to actions (e.g., a chess move leads to checkmate or loss). Statistical regularities dominate, enabling pattern-based strategies.
\textit{Wicked} learning environments, on the other hand, are characterized by dynamic rules, sparse or ambiguous feedback, and novelty. Examples include entrepreneurship, healthcare diagnostics, and customer engagement. Outcomes may emerge long after actions (e.g., a marketing campaign's impact on brand loyalty), and underlying data distributions shift over time (e.g., evolving consumer preferences). Hogarth argued that human intuition excels in kind environments but falters in wicked ones, necessitating deliberate analytical strategies. 

LLMs are an embodiment of human intuition, and exhibited through language. LLMs excel in kind environments due to transformer-based pattern matching. For example: given an error message, an LLM predicts fixes to the code by correlating with repair patterns in training data. Even in these remarkably stable and well-known environments, LLMs still overfit to training distributions, causing them to fail on edge cases outside their corpus~\cite{Aissi2024}.

LLMs have no way to successfully navigate wicked learning environments. For example, an LLM chat bot trained on retail dialogues may handle queries like ``find navy blue shirts'' (records retrieval is a kind learning environment), but fail when a user abruptly exits a session after receiving recommendations.  LLMs process each interaction as an independent sequence, with no persistent memory of prior sessions. The user's exit could signal dissatisfaction, distraction, or indecision---a wicked feedback signal the LLM cannot interpret. LLMs assume a fixed data distribution based on their training data, but wicked environments require adaptation to distributions outside of the training data.

It might be argued that LLMs could handle wicked learning environments via fine-tuning, but
\begin{enumerate*}[label=(\roman*)]
    \item fine-tuning on new data overwrites prior weights, degrading performance on original tasks (catastrophic forgetting), and
    \item continuous fine-tuning is impractical and in most cases financially infeasible for real-time adaptation.
\end{enumerate*}
The ``kind vs. wicked'' dichotomy (which, of course, is more of a continuous spectrum than a set of hard-and-fast categories) mandates distinct technical strategies: in kind learning environments, we can leverage LLMs' procedural strengths but in wicked learning environments, agents need semantic-associative memory to be able to adapt appropriately. 
\section{Augmenting LLMs with Modular Semantic-Associative Systems}
To enable robust decision-making in wicked environments, we propose a modular architecture where agentic \textit{learners} (semantic-associative systems) operate independently of agentic \textit{actors} (LLMs). This separation ensures specialized cognitive capabilities: learners focus on adaptive reasoning via reinforcement learning (RL)~\cite{Sutton1998}, while actors handle procedural execution. 

LLMs act as context-bound agents, limited to processing the inputs they're given. Their architecture prevents three key capabilities: 
\begin{enumerate*}[label=(\roman*)]
    \item retaining distilled learnings across sessions,
    \item dynamically associating actions with outcomes outside of explicit user feedback, or
    \item autonomously expanding their context.
\end{enumerate*}
These aren't scaling problems---they're inherent to transformers' stateless design. However, to successfully navigate wicked learning environments, agentic learners are needed, capable of incrementally building context by exploring and linking semantic categories.

Agentic learners input actions---or, more precisely, they input a stream of user interactions with different actions and, via model-free RL and explore-exploit mechanisms like Thompson sampling~\cite{Thompson1933}, they construct contextual metadata that can then be used to select next best actions, optionally generating those actions by passing that metadata to an agentic actor.
An agentic learner can operate on multiple action sets at the same time.

Customer engagement---the ongoing process of influencing user behavior through personalized interactions---is a good example of a wicked learning environment. Success requires adapting to shifting preferences, interpreting ambiguous signals (e.g., a purchase might indicate satisfaction or mere acquiescence), and evolving strategies without clear rules. Traditional LLMs fail here because they treat each interaction as independent, unable to link outcomes to semantic distillations, and therefore unable to build a reasonably tidy and memory-efficient context across sessions.  
Our proposed architecture separates adaptation from execution. An independent learning module handles longitudinal reasoning: tracking which strategies work for specific users, associating causes (e.g., communication timing and frequency, channel, recommendation approaches, copy elements, etc.) with effects (e.g., user engagement metrics), and refining its understanding as new data arrives. The LLM then translates these distilled insights into natural language, focusing solely on coherent response generation.  

This division of labor plays to each component's strengths. The learner operates on slightly slower timescales, filtering noise from sparse feedback, while the LLM handles real-time communication. Crucially, the system doesn't seek deterministic answers. Rather, it maintains competing hypotheses about user preferences, updating them probabilistically as evidence accumulates. This mirrors how humans navigate uncertainty, but with the scalability required for automated systems.  

\subsection{Integration with LLM Actors}
The agentic learner generates contextual metadata by compiling contextual information. This metadata encapsulates the learner's distilled understanding of the user's preferences and the current context, serving as a bridge between adaptive reasoning and procedural execution. Once the context vector is constructed, it is passed to the LLM actor as a prefix to the user's prompt. The LLM then generates a response conditioned on this augmented input. This integration ensures that the LLM's procedural strengths---fluency, coherence, and stylistic adaptability---are guided by the learner's semantic and associative insights.

Agentic learners can be said to have a cold-start problem, though this is arguably not a flaw, but a design feature of agentic learners operating in wicked environments. Procedures like sliding-window Thompson sampling~\cite{Trovo2020} inherently address this via their explore-exploit balance. Early interactions will prioritize broad exploration (e.g., testing all value propositions equally). In wicked environments, user preferences will shift, and the resulting lack of rewards for previously-rewarded semantic categories will shift and flatten the underlying distributions for those categories, causing an agentic learner to automatically reallocate exploration bandwidth.

Agentic learners embrace uncertainty: they are not tasked with finding a ``right'' answer (which rarely exists in wicked learning environments), but with placing adaptively informed bets. In other words, they are designed for causal decision-making rather than causal estimation~\cite{Fernández-Loría2021}.

\section{Modular Architectures for Cognitive Specialization}
The human brain has several different definitions of cognition, each backed by different mechanisms, whereas LLMs have only one definition of cognition, backed by only one mechanism. This raises serious doubts about the ability of LLMs to serve as a foundation for broader ``Artificial General Intelligence'' initiatives, or even for the more modest goal of building systems that are capable of acting agentically in messy and complex real-world contexts.

Current efforts to expand LLMs into multi-modal systems often conflate mechanical diversity---such as processing images, text, or audio---with cognitive specialization. While multi-modal systems enhance the range of input types an agent can handle, they do not address the fundamental need for specialized cognitive subsystems. True autonomy requires a modular architecture where distinct components are optimized for specific cognitive tasks. For example: 
\begin{itemize}
    \item A semantic module manages structured knowledge by organizing learned actions and concepts into abstract, generalizable representations---similar to how human semantic memory encodes facts and rules detached from specific experiences~\cite{Tulving1972};
    \item An associative module links experiences by forming and retrieving relationships between co-occurring states and actions---a process analogous to associative binding in human cognition, where elements of experience are connected through repeated or meaningful co-activation~\cite{Ranganath2012}.
    \item A procedural module uses learned semantic associations as context to generate coherent, human-readable responses.  
\end{itemize}
By decoupling these functions, the system avoids the pitfalls of monolithic architectures, where a single model is forced to handle tasks for which it is ill-suited. While this kind of modularity introduces interface challenges---such as coordinating between components and managing error propagation---these are deliberate trade-offs. A modular design offers clearer control boundaries, improves interpretability of agent behavior, and allows individual modules to be retrained or upgraded without overhauling the entire system.

\section{Related Work}

Cognitive architectures such as \textit{ACT-R} and \textit{SPAUN} take a more explicit stance on modularity. \textit{ACT-R} models human cognition as an interplay of symbolic modules tied to psychological theory~\cite{anderson1996actr}, while \textit{SPAUN} integrates spiking neural networks with a symbolic architecture to simulate a wide range of cognitive tasks~\cite{eliasmith2012spaun}. These systems prioritize biological plausibility and general cognitive competence but typically rely on hand-engineered structures, which limits their scalability and adaptability to novel environments.

More recent approaches emphasize general-purpose control across multiple modalities. LeCun's Joint Embedding Predictive Architecture proposes a framework where world modeling and goal conditioning are learned through predictive embeddings~\cite{lecun2022jeppa}, while \textit{Gato} frames intelligence as a unified sequence modeling problem trained across diverse tasks and input modalities~\cite{reed2022gato}. Both aim at generalization through architectural homogeneity and large-scale pretraining, but offer limited support for dynamic modular composition or interpretability during inference.

Our approach departs from these lines of work in several ways. While we embrace sub-symbolic representation and end-to-end learning, we reject architectural monoliths in favor of dynamic modularity. Rather than aiming for universality or cognitive emulation, we build systems from composable components that can flexibly exchange context and specialize within wicked, real-world environments. Modularity is not merely a structural convenience—it is a mechanism for ongoing adaptation, contextual transparency, and grounded control. This allows our system to balance generalization with interpretation and reuse, even in situations that defy fixed task boundaries or static state representations.

The design of agent architectures must prioritize the learning environment in which they operate. In kind environments (characterized by stable rules, repetitive patterns, and clear feedback), LLMs can function effectively as standalone agents. However, in wicked environments (where rules are dynamic, feedback is ambiguous, and novelty is the norm), hybrid systems are essential. This division of labor ensures that the system can adapt to the unpredictability of wicked environments while maintaining the procedural strengths of LLMs.  

\section{Conclusion}
Real-world decision-making requires cognitive diversity beyond procedural memory. To advance autonomous agents, we propose three key principles:
\begin{enumerate}
    \item Decoupling Cognitive Modules. LLMs should act as components within modular architectures, not central controllers. This separation allows each module to specialize in its respective cognitive function, whether procedural, associative, or semantic.  
    \item Rigorous Environment Classification. Both application and data systems must be designed with the learning environment in mind. Kind environments may require only procedural capabilities, while wicked environments demand hybrid architectures that integrate associative and semantic reasoning.  
    \item Investment in Associative and Semantic Systems. Prioritize research into neural-symbolic architectures, sparse memory models, and other frameworks that enable explicit reasoning and adaptive learning.  
\end{enumerate}
By embracing these principles, we can develop agents that complement human ingenuity in uncertainty—not merely replicate procedural expertise in structured domains. This shift from monolithic to modular architectures represents a necessary evolution in AI design, one that acknowledges the complexity of real-world decision-making and the limitations of current approaches.

\bibliographystyle{ACM-Reference-Format}
\bibliography{bibliography}

\end{document}